\documentclass{article}
\usepackage[colorlinks,linkcolor=black,citecolor=black,urlcolor=black]{hyperref}
\usepackage{url}

\usepackage{graphicx}
\usepackage{amsmath,amssymb,amsfonts,amstext,amsthm,mathrsfs}
\usepackage{thmtools}
\usepackage{mathtools}
\usepackage{thm-restate}
\usepackage{latexsym,epsf,psfrag,epsfig,epstopdf}
\graphicspath{{./figs/}}
\usepackage[sort&compress]{natbib}
\usepackage{cleveref}
\usepackage{cite}
\usepackage{dsfont}

\usepackage{algorithm}
\usepackage{algorithmic}
\usepackage{caption}
\usepackage{subcaption}
\usepackage{color}
\usepackage{enumitem}

\newtheorem{definition}{Definition}

\newtheorem{thm}{Theorem}

\newtheorem{assum}{Assumption}

\usepackage{xspace}

\newcommand{\eg}{{e.g.}\xspace}
\newcommand{\ie}{{i.e.}\xspace}

\newcommand{\gb}{\mathbf{g}}
\newcommand{\eb}{\mathbf{e}}
\newcommand{\xb}{\mathbf{x}}

\newcommand{\yb}{\mathbf{y}}
\newcommand{\zb}{\mathbf{z}}

\newcommand{\ub}{\mathbf{u}}
\newcommand{\vb}{\mathbf{v}}
\newcommand{\RR}{\mathds{R}}
\newcommand{\Ebb}{\mathbb{E}}
\newcommand{\dist}{\mathrm{dist}}
\newcommand{\prox}[1]{\mathrm{prox}_{#1}}

\newcommand{\dom}{\mathop{\mathrm{dom}}}
\newcommand{\zero}{\mathbf{0}}

\newcommand{\argmin}{\mathop{\mathrm{argmin}}}
\newcommand{\KL}{K\L\xspace}
\newcommand{\inner}[2]{\langle #1, #2 \rangle}

\usepackage[accepted]{icml2017}

\icmltitlerunning{Convergence Analysis of Proximal Gradient with Momentum for Nonconvex Optimization}

\begin{document}

\twocolumn[
\icmltitle{Convergence Analysis of Proximal Gradient with Momentum for Nonconvex Optimization}

\begin{icmlauthorlist}
\icmlauthor{Qunwei Li}{to}
\icmlauthor{Yi Zhou}{to}
\icmlauthor{Yingbin Liang}{to}
\icmlauthor{Pramod K. Varshney}{to}
\end{icmlauthorlist}

\icmlaffiliation{to}{Syracuse University, NY, USA}

\icmlcorrespondingauthor{Qunwei Li}{qli33@syr.edu}

\icmlkeywords{boring formatting information, machine learning, ICML}

\vskip 0.3in
]

\printAffiliationsAndNotice{} % otherwise use the standard text.
%\footnotetext{hi}

\begin{abstract}
In many modern machine learning applications, structures of underlying mathematical models often yield nonconvex optimization problems. Due to the intractability of nonconvexity, there is a rising need to develop efficient methods for solving general nonconvex problems with certain performance guarantee. In this work, we investigate the accelerated proximal gradient method for nonconvex programming (APGnc) \cite{Yao_2016}. The method compares between a usual proximal gradient step and a linear extrapolation step, and accepts the one that has a lower function value to achieve a monotonic decrease. In specific, under a general nonsmooth and nonconvex setting, we provide a rigorous argument to show that the limit points of the sequence generated by APGnc are critical points of the objective function. Then, by exploiting the Kurdyka-{\L}ojasiewicz (\KL) property for a broad class of functions, we establish the linear and sub-linear convergence rates of the function value sequence generated by APGnc. We further propose a stochastic variance reduced APGnc (SVRG-APGnc), and establish its linear convergence under a special case of the \KL property. We also extend the analysis to the inexact version of these methods and develop an adaptive momentum strategy that improves the numerical performance.
\end{abstract}

%!TEX root = paper.tex

\section{Introduction}\label{intro}
%In modern data science, including machine learning, statistical learning, data mining, and signal processing, etc, a common and popular way to formulate the problems of interest is to formalize them into the optimization framework. In particular,
Many problems in machine learning, data mining, and signal processing can be formulated as the following composite minimization problem
\begin{align*}
	\tag{P} \min_{\xb \in \RR^d} ~F(\xb) = f(\xb) + g(\xb).
\end{align*}
Typically, $f: \RR^d \to \RR$ captures the loss of data fitting and can be written as $f = \frac{1}{n}\sum_{l=1}^{n} f_l$ with each $f_l$ corresponding to the loss of one sample. The second term $g: \RR^d \to \RR$ is the regularizer that promotes desired structures on the solution based on prior knowledge of the problem.

%Generally, problems of \textbf{(P)} can be roughly divided into two categories: convex problems and non-convex problems, which distinguishes between their algorithm tractabilities.

%Convex problems, essentially, can be approximated below globally by its first order Taylor expansion. The key property that makes convex functions attractive is that the local minimum are also global minimum, and this allows one to track the global convergence behavior of many popular algorithms. Hence,
In practice, many problems of \textbf{(P)} are formulated, either naturally or intensionally, into a convex model to guarantee the tractability of algorithms.
%Some concrete examples include the SVM (support vector machine), $l_1$ Lasso, sparse logistic regression, ridge regression, matrix completion with nuclear norm, etc.
%In particular, a lot of such convex examples share a common structure that the data fitting part $f$ has a Lipschitz continuous gradient (also referred to as smooth).
In particular, such convex problems can be efficiently minimized by many first-order algorithms, among which the accelerated proximal gradient (APG) method (also referred to as FISTA \cite{FISTA}) is proven to be the best for minimizing such class of convex functions. We present one of its basic forms in \Cref{alg: APG}.
\begin{algorithm}[tb]
	\caption{APG}\label{alg: APG}
	\begin{algorithmic}
		\STATE {\bfseries Input:} $\yb_1 = \xb_1 = \xb_0, t_1 = 1, t_0 = 0, \eta < \frac{1}{L}$.
		\FOR{$k=1, 2, \cdots$}
		\STATE $\yb_k = \xb_k + \frac{t_{k-1} - 1}{t_k}(\xb_k - \xb_{k-1})$.
		\STATE $\xb_{k+1} = \prox{\eta g} (\yb_k - \eta \nabla f(\yb_k))$.
		\STATE $t_{k+1} = \frac{\sqrt{4t_k^2 + 1} + 1}{2}$.
		\ENDFOR
	\end{algorithmic}
\end{algorithm}
Compared to the usual proximal gradient step, the APG algorithm takes an extra linear extrapolation step for acceleration. It has been shown \cite{FISTA} that the APG method reduces the function value gap at a rate of $\mathcal{O}(1/k^2)$ where $k$ denotes the number of iterations. This convergence rate meets the theoretical lower bound of first-order gradient methods for minimizing smooth convex functions. The reader can refer to \cite{Tseng_2010} for other variants of APG.

%Although convex problems are tractable and can be globally minimized, the restriction of convexity often violates the nature of many applications. Thus, one may lose the accuracy of the mathematical model in exchange for the tractability of the algorithms, and this is unbearable for many applications. For example, in sparsity promoting problems, the non-convex $l_0$ quasi norm is more natural than the convex $l_1$ norm, while in low rank problems rank is more accurate than its convex relaxation nuclear norm. Other examples that naturally take non-convex formulations include phase retrieval, matrix factorization, low rank matrix completion, etc. Thus, it is desirable to develop an corresponding APG-like method for these non-convex problems, and provide a formal convergence guarantee.

\begin{algorithm}
		\caption{Monotone APG (mAPG)}\label{alg: mAPG}
		\begin{algorithmic}
			\STATE {\bfseries Input:} $\yb_1 = \xb_1 = \xb_0, t_1 = 1, t_0 = 0, \eta < \frac{1}{L}$.
			\FOR{$k=1, 2, \cdots$}
			\STATE $\yb_k = \xb_k + \frac{t_{k-1}}{t_k}(\zb_k - \xb_{k}) + \frac{t_{k-1} - 1}{t_k}(\xb_k - \xb_{k-1})$.
			\STATE $\zb_{k+1} = \prox{\eta g} (\yb_k - \eta \nabla f(\yb_k))$.
			\STATE $\vb_{k+1} = \prox{\eta g} (\xb_k - \eta \nabla f(\xb_k))$.
			\STATE $t_{k+1} = \frac{\sqrt{4t_k^2 + 1} + 1}{2}$.
			\IF{$F(\zb_{k+1}) \le F(\vb_{k+1})$}
			\STATE $\xb_{k+1} = \zb_{k+1},$
			\ELSIF{$F(\vb_{k+1}) \le F(\zb_{k+1})$}
			\STATE $\xb_{k+1} = \vb_{k+1}$.
			\ENDIF
			\ENDFOR
		\end{algorithmic}
	\end{algorithm}

 	\begin{algorithm}[tb]
 		\caption{APG non-convex problem (APGnc)}
 		\label{alg: APGnc}
 		\begin{algorithmic}
 			\STATE {\bfseries Input:} $\yb_1 = \xb_0, \beta_k = \frac{k}{k+3}, \eta < \frac{1}{L}$.
 			\FOR{$k=1, 2, \cdots$}
 			\STATE $\xb_k = \prox{\eta g} (\yb_k - \eta \nabla f(\yb_k))$.
 			\STATE $\vb_k = \xb_k + \beta_k(\xb_k - \xb_{k-1})$.
 			\IF{$F(\xb_k) \le F(\vb_k)$}
 			\STATE $\yb_{k+1} = \xb_k,$
 			\ELSIF{$F(\vb_k) \le F(\xb_k)$}
 			\STATE $\yb_{k+1} = \vb_k$.
 			\ENDIF
 			\ENDFOR
 		\end{algorithmic}
 	\end{algorithm}

Although convex problems are tractable and can be globally minimized, many applications naturally require to solve nonconvex optimization problems of \textbf{(P)}. Recently, several variants of the APG method have been proposed for nonconvex problems, and two major ones are presented in \Cref{alg: mAPG} and \Cref{alg: APGnc}, respectively. The major difference to the original APG is that the modified methods only accept the new iterate when the corresponding function value is sufficiently decreased, which leads to a more stable convergence behavior. In particular, \cite{Lihuan_2015} analyzed mAPG (\Cref{alg: mAPG}) by exploiting the Kurdyka-{\L}ojasiewicz (\KL) property, which is a local geometrical structure very generally held by a large class of nonconvex objective functions, and has been successfully exploited to characterize the asymptotic convergence behavior of many first order methods. It was shown in \cite{Lihuan_2015} that mAPG achieves the $\mathcal{O}(1/k^2)$ convergence rate for convex problems of \textbf{(P)}, and converges to a critical point at sublinear and linear rates under different cases of the KL property for nonconvex problems. Despite the desirable convergence rate, mAPG requires two proximal steps, which doubles the computational complexity of the original APG. In comparison, the APGnc (\Cref{alg: APGnc}) requires only one proximal step, and hence computes faster than mAPG in each iteration. However, the analysis of APGnc in \cite{Yao_2016} does not exploit the KL property and no convergence rate of the function value is established. Hence, there is still no formal theoretical comparison of the overall performance (which depends on both computation per iteration and convergence rate) between mAPG and APGnc. It is unclear whether the computational saving per iteration in APGnc is at the cost of lower convergence rate.

The goal of this paper is to provide a comprehensive analysis of the APGnc algorithm under the KL framework, thus establishing a rigorous comparison between mAPG and APGnc and formally justifying the overall advantage of APGnc.

\subsection{Main Contributions}
%Thm 1, \cite{Yao_2016} only fixed point while our critical point (in the sense of limiting subdiff);  \cite{Lihuan_2015}, simpler method achieves the same key properties; inexact, two sequences compare to original inexact \KL, proximal map error.

This paper provides the convergence analysis of APGnc type algorithms for {\em nonconvex} problems of \textbf{(P)} under the KL framework as well as the inexact situation. We also study the stochastic variance reduced APGnc algorithm and its inexact situation. Our analysis requires novel technical treatments to exploit the \KL property due to the joint appearance of the following ingredients in the algorithms including momentum terms, inexact errors, and stochastic variance reduced gradients. Our contributions are summarized as follows.

For APGnc applied to nonconvex problems of \textbf{(P)}, we show that the limit points of the sequences generated by APGnc are critical points of the objective function. Then, by exploiting different cases of the Kurdyka-{\L}ojasiewicz property of the objective function, we establish the linear and sub-linear convergence rates of the function value sequence generated by APGnc. Our results formally show that APGnc (with one proximal map per iteration) achieves the same convergence properties as well as the convergence rates as mAPG (with two proximal maps per iteration) for nonconvex problems, thus establishing its overall computational advantage.

We further propose an APGnc$^+$ algorithm, which is an improved version of APGnc by adapting the momentum stepsize (see \Cref{alg: APGam}), and shares the same theoretical convergence rate as APGnc but numerically performs better than APGnc.
%with appropriately adapted momentum, which achieves the same order-level convergence rate as APGnc but numerically performs better than APGnc.

Furthermore, we study the inexact APGnc in which the computation of the gradient and the proximal mapping may have errors. We show that the algorithm still achieves the convergence rate at the same order as the exact case as long as the inexactness is properly controlled. We also explicitly characterize the impact of errors on the constant factors that affect the convergence rate.

To facilitate the solution to large-scale optimization problems, we study the stochastic variance reduced APGnc (SVRG-APGnc), and show that such an algorithm achieves linear convergence rate under a certain case of the \KL property. We further analyze the inexact SVRG-APGnc and show that it also achieves the linear convergence under the same \KL property as long as the error in the proximal mapping is bounded properly. This is the first analysis of the SVRG proximal algorithm with {\em momentum} that exploits the \KL structure to establish linear convergence rate for nonconvex programming.

Our numerical results further corroborate the theoretic analysis. We demonstrate that APGnc/APGnc$^+$ outperforms APG and mAPG for nonconvex problems in both exact and inexact cases, and in both deterministic and stochastic variants of the algorithms. Furthermore, APGnc$^+$ outperforms APGnc due to properly chosen momentum stepsize.

%Novelty in technical proof: (1) no much difference in proof of APGnc from mAPG under KL; (2) (For deterministic and convex $g$, we characterize the inexactness of the proximal map) special technical treatments in inexact case; (3) special technical treatment in SVRG APGnc under PL; difference from CMU paper. (We use \KL for composite functions, they PL for single function)

\subsection{Comparison to Related Work}
\textbf{APG algorithms:} The original accelerated gradient method for minimizing a single smooth convex function dates back to \cite{Nesterov_1983}, and is further extended as APG in the composite minimization framework in \cite{FISTA,Tseng_2010}. While these APG variants generate a sequence of function values that may oscillate, \cite{Beck_2009} proposed another variant of APG that generates a non-increasing sequence of function values. Then, \cite{Lihuan_2015} further proposed an mAPG that generates a sufficiently decreasing sequence of function values, and established the asymptotic convergence rates under the \KL property. Recently, \cite{Yao_2016} proposed APGnc, which is a more efficient version of APG for nonconvex problems, but the analysis only characterizes fixed points and did not exploit the \KL property to characterize the convegence rate. Our study establishes the convergence rate analysis of APGnc under the \KL property.

\textbf{Nonconvex optimization under \KL:} The \KL property \cite{KL_extend_nonsmooth} is an extension of the {\L}ojasiewicz gradient inequality \cite{Lojas_1965} to the nonsmooth case. Many first-order descent methods, under the \KL property, can be shown to converge to a critical point \cite{KL_gene_rate,prox_alter_linear,KL_bound_sequence} with different types of asymptotic convergence rates. \cite{Lihuan_2015} and our paper focuses on the first-order algorithms with {\em momentum}, and respectively analyze mAPG and APGnc by exploiting the \KL property.

\textbf{Inexact algorithms under \KL:} \cite{attouch_descent_semi,Frankel2015} studied the inexact proximal algorithm under the \KL property. This paper studies the inexact proximal algorithm with momentum (i.e., APGnc) under the \KL property. While \cite{Yao_2016} also studied the inexact APGnc, the analysis did not exploit the \KL property to characterize the convergence rate.
%does not specify the inexactness of the proximal map.

\textbf{Nonconvex SVRG:}
SVRG was first proposed in \cite{johnson2013accelerating}, to accelerate the stochastic gradient method for strongly convex objective functions, and was studied for the convex case in \cite{allen2016improved}. Recently, SVRG was further studied for smooth nonconvex optimization in \citet{reddi2016stochastic}. Then in \cite{NIPS2016_6116}, the proximal SVRG was proposed and studied for nonsmooth and nonconvex optimization. Our paper further incorporates SVRG for the proximal gradient with {\em momentum} in the nonconvex case. Furthermore, we exploit a certain \KL property in our analysis that is very different from the PL property exploited in \cite{reddi2016stochastic}, and requires special technical treatment in convergence analysis.

%Hinged upon variance reduction techniques, stochastic method SVRG has achieved remarkable progress for smooth convex and nonconvex instances of problem ${\mathbf{P}}$ for large scale optimization \cite{johnson2013accelerating,reddi2016stochastic}. Proximal version of SVRG is first studied in \cite{shamir2014stochastic}, and is then followed up by \cite{xiao2014proximal}. The convergence rates of proximal SVRG for strongly-convex and nonconvex cases are developed respectively in \cite{nitanda2014stochastic} and \cite{NIPS2016_6116}.

%To exploit the \KL property in nonconvex optimization with proximal SVRG, we construct a new version of proximal SVRG that monotone iterates can be generated, where momentum is also considered for acceleration.

%!TEX root = paper.tex

\section{Preliminaries and Assumptions}\label{sec: Pre}

In this section, we first introduce some technical definitions that are useful later on, and then describe the assumptions on the problem \textbf{(P)} that we take in this paper.

Throughout this section, $h:\RR^d \to (-\infty, +\infty]$ is an extended real-valued function that is proper, \ie, its domain $\dom h := \{ \xb \in \RR^d: h(\xb) < \infty \}$ is nonempty,  and is closed, \ie, its sublevel sets $\{ \xb\in \RR^d: h(\xb) \leq \alpha \}$ are closed for all $\alpha\in\RR$. Note that a proper and closed function $h$  can be nonsmooth and nonconvex, hence we consider the following generalized notion of derivative.
\begin{definition}[Subdifferential, \cite{vari_ana}]
\label{def:sub}
The Frech\'et subdifferential $\hat\partial h$ of $h$ at $\xb\in\dom h$ is the set of $\ub \in \RR^d$ such that
\begin{align*}
\liminf_{\zb\neq\xb, \zb\to\xb} \tfrac{h(\zb) - h(\xb) - \ub^\top(\zb-\xb)}{\|\zb-\xb\|} \geq 0,
\end{align*}
while the (limiting) subdifferential $\partial h$ at $\xb\in\dom h$ is the graphical closure of $\hat\partial h$:
\begin{align*}
\{ \ub: \exists (\xb^k, h(\xb^k)) \to (\xb, h(\xb)), \hat{\partial} h(\xb^k) \ni \ub^k \to\ub \}.
\end{align*}
\end{definition}
In particular, this generalized derivative reduceds to $\nabla h$ when $h$ is continuously differentiable, and reduces to the usual subdifferential when $h$ is convex.
\begin{definition}[Critical point]
A point $\xb\in \RR^d$ is a critical point of $h$ iff $\zero \in \partial h(\xb) $.
\end{definition}

\begin{definition}[Distance]\label{def: distance}
The distance of a point $\xb \in \RR^d$ to a closed set $\Omega \subseteq \RR^d$ is defined as:
\begin{align}
\textstyle
\dist_{\Omega}(\xb) := \min_{\yb\in\Omega} \|\yb-\xb\|.
\end{align}
%and the metric projection onto $\Omega$ is:
%\begin{align}
%\textstyle
%\proj{\Omega}(\xb) := \argmin_{\yb\in\Omega} \|\yb-\xb\|.
%\end{align}
\end{definition}
%Note that $\proj{\Omega}$ is always a singleton iff $\Omega$ is convex.

\begin{definition}[Proximal map, \eg \cite{vari_ana}]
\label{def:prox}
The proximal map of a point $\xb \in \RR^d$ under a proper and closed function $h$ with parameter $\eta > 0$ is defined as:
\begin{align}
\textstyle
\prox{\eta h}(\xb) := \argmin_{\zb} h(\zb) + \tfrac{1}{2\eta}\|\zb - \xb\|^2,
\end{align}
where $\|\cdot\|$ is the Euclidean $l_2$ norm.
\end{definition}
We note that when $h$ is convex, the corresponding proximal map is the minimizer of a strongly convex function, \ie, a singleton. But for nonconvex $h$, the proximal map can be set-valued, in which case $\prox{\eta h}(\xb)$ stands for an arbitrary element from that set. The proximal map is a popular tool to handle the nonsmooth part of the objective function, and is the key component of proximal-like algorithms \cite{FISTA,prox_alter_linear}.
\begin{definition}[Uniformized \KL property, {\cite{prox_alter_linear}}]
	\label{thm:KL}
	Function $h$ is said to satisfy the uniformized \KL property if for every compact set $\Omega\subset \mathrm{dom}h$ on which $h$ is constant, there exist $\varepsilon, \lambda >0$ such that for all $\bar{\xb} \in \Omega$ and all $\xb\in \{\xb\in \RR^d : \dist_\Omega(\xb)<\varepsilon\}\cap [\xb: h(\bar{\xb}) < h(\xb) <h(\bar{\xb}) + \lambda]$, one has
	\begin{align}
	\label{KLineq}
	\varphi' \left(h(\mathbf{x}) - h(\bar{\mathbf{x}})\right) \cdot \dist_{\partial h(\xb)}(\zero) \ge 1,
	\end{align}
	where the function $\varphi: [0,\lambda) \to \RR_+$ takes the form $\varphi(t) = \frac{c}{\theta} t^\theta$ for some constants $c>0, \theta\in (0,1]$.
\end{definition}
The above definition is a modified version of the original \KL property \cite{initial_KL, BolteDLM10}, and is more convenient for our analysis later. The \KL property is a generalization of the {\L}ojasiewicz gradient inequality to nonsmooth functions \cite{KL_extend_nonsmooth}, and it is a powerful tool to analyze a class of first-order descent algorithms \cite{KL_gene_rate,prox_alter_linear,KL_bound_sequence}.
In particular, the class of semi-algebraic functions satisfy the above \KL property. This function class covers most objective functions in real applications, for instance, all $l_p$ where $p\ge 0$ and is rational, real polynomials, rank, etc. For a more detailed discussion and a list of examples of \KL functions, see \cite{prox_alter_linear} and \cite{KL_bound_sequence}.

We adopt the following assumptions on the problem \textbf{(P)} in this paper.
\begin{assum}\label{assum:func}
	Regarding the functions $f, g$ (and $F=f+g$) in \textbf{(P)}
	\begin{enumerate}[leftmargin=*,topsep=0pt,noitemsep]
		\item They are proper and lower semicontinous; $\inf_{\xb\in \RR^d} F(\xb) > -\infty$; the sublevel set $\{ \xb\in \RR^d: F(\xb) \leq \alpha \}$ is bounded for all $\alpha \in \RR$;
		\item They satisfy the uniformized \KL property;
		\item Function $f$ is continuously differentiable and the gradient $\nabla f$ is $L$-Lipschitz continuous.
	\end{enumerate}
\end{assum}
Note that the sublevel set of $F$ is bounded when either $f$ or $g$ has bounded sublevel set, \ie, $f(\xb)~ \text{or}~ g(\xb) \to +\infty$ as $\|\xb\| \to +\infty$.
Of course, we do \emph{not} assume convexity on either $f$ or $g$, and the \KL property serves as an alternative in this general setting.

%!TEX root = paper.tex

\section{Main Results}\label{sec: form_analy}

In this section, we provide our main results on the convergence analysis of APGnc and SVRG-APGnc as well as inexact variants of these algorithms. All proofs of the theorems are provided in supplemental materials.

\subsection{Convergence Analysis}
In this subsection, we characterize the convergence of APGnc. Our first result characterizes the behavior of the limit points of the sequence generated by APGnc.
\begin{thm}\label{thm: limit_point}
	Let \Cref{assum:func}.\{1,3\} hold for the problem \textbf{(P)}. Then with stepsize $\eta < \frac{1}{L}$, the sequence $\{\xb_k\}$ generated by APGnc satisfies
	\begin{enumerate}[leftmargin=*,topsep=0pt,noitemsep]
		\item $\{\xb_k\}$ is a bounded seuqence;
		\item The set of limit points $\Omega$ of $\{\xb_k\}$ forms a compact set, on which the objective function $F$ is constant;
		\item All elements of $\Omega$ are critical points of $F$.
	\end{enumerate}
\end{thm}
\Cref{thm: limit_point} states that the sequence $\{\xb_k\}$ generated by APGnc eventually approaches a compact set (\ie, a closed and bounded set in $\RR^d$) of critical points, and the objective function remains constant on it. Here, approaching critical points establishes the first step for solving general nonconvex problems. Moreover, the compact set $\Omega$ meets the requirements of the uniform \KL property, and hence provides a seed to exploit the \KL property around it. Next, we further utilize the \KL property to establish the asymptotic convergence rate for APGnc. In the following theorem, $\theta$ is the parameter in the uniformized \KL property via the function $\varphi$ that takes the form $\varphi(t) = \frac{c}{\theta} t^\theta$ for some $c>0, \theta \in (0,1]$.
\begin{thm}\label{thm: KL_rate}
	Let \Cref{assum:func}.\{1,2,3\} hold for the problem \textbf{(P)}. Let $F(\xb) \equiv F^*$ for all $\xb \in \Omega$ (the set of limit points), and denote $r_k := F(\xb_k) - F^*$. Then with stepsize $\eta < \frac{1}{L}$, the sequence $\{r_k\}$ satisfies for $k_0$ large enough
	\begin{enumerate}[leftmargin=*,topsep=0pt,noitemsep]
		\item If $\theta = 1$, then $r_k$ reduces to zero in finite steps;
		\item If $\theta \in [\frac{1}{2}, 1)$, then $r_k \le (\frac{c^2d_1}{1+c^2d_1})^{k-k_0} r_{k_0}$;
		\item If $\theta \in (0, \frac{1}{2})$, then $r_k \le (\frac{c}{(k-k_0)d_2 (1-2\theta)})^{\frac{1}{1-2\theta}}$,
	\end{enumerate}
	where $d_1 = (\frac{1}{\eta} + L)^2/ (\frac{1}{2\eta} - \frac{L}{2})$ and $d_2 = \min \{\frac{1}{2cd_1}, \frac{c}{1-2\theta}(2^{\frac{2\theta - 1}{2\theta - 2}} - 1)r_{k_0}^{2\theta - 1} \}$.
\end{thm}
%Theorem \Cref{thm: KL_rate} are typical in the convergence analysis with \KL property, and it
\Cref{thm: KL_rate} characterizes three types of convergence behaviors of APGnc, depending on $\theta$ that parameterizes the \KL property that the objective function satisfies. An illustrative example for the first kind ($\theta = 1$) can take a form similar to $F(x) = |x|$ for $x\in\RR$ around the critical points. The function is `sharp' around its critical point $x=0$ and thus the iterates slide down quickly onto it within finite steps. For the second kind ($\theta \in [\frac{1}{2}, 1)$), example functions can take a form similar to $F(x) = x^2$ around the critical points. That is, the function is strongly convex-like and hence the convergence rate is typically linear. Lastly, functions of the third kind are `flat' around its critical points and thus the convergence is slowed down to sub-linear rate. We note that characterizing the value of $\theta$ for a given function is a highly non-trivial problem that takes much independent effort \cite{Li-Kei-2016,kurdyka2011}. Nevertheless, \KL property provides a general picture of the asymptotic convergence behaviors of APGnc.

\subsection{APGnc with Adaptive Momentum}
The original APGnc sets the momentum parameter $\beta_k = \frac{k}{k+3}$, which can be theoretically justified only for convex problems. We here propose an alternative choice of the momentum stepsize that is more intuitive for nonconvex problems, and refer to the resulting algorithm as APGnc$^+$ (See \Cref{alg: APGam}). The idea is to enlarge the momentum $\beta$ to further exploit the opportunity of acceleration when the extrapolation step $\vb_k$ achieves a lower function value. Since the proofs of \Cref{thm: limit_point} and \Cref{thm: KL_rate} do not depend on the exact value of the momentum stepsize, APGnc and APGnc$^+$ have the same order-level convergence rate. However, we show in \Cref{sec: exp} that APGnc$^+$ improves upon APGnc numerically.
\begin{algorithm}[tb]
	\caption{APGnc with adaptive momentum (APGnc$^+$)}
	\label{alg: APGam}
	\begin{algorithmic}
		\STATE {\bfseries Input:} $\yb_1 = \xb_0, \beta ,t \in (0,1), \eta < \frac{1}{L}$.
		\FOR{$k=1, 2, \cdots$}
		\STATE $\xb_k = \prox{\eta g} (\yb_k - \eta \nabla f(\yb_k))$.
		\STATE $\vb_k = \xb_k + \beta_k(\xb_k - \xb_{k-1})$.
		\IF{$F(\xb_k) \le F(\vb_k)$}
		\STATE $\yb_{k+1} = \xb_k, \beta \gets t\beta.$
		\ELSIF{$F(\vb_k) \le F(\xb_k)$}
		\STATE $\yb_{k+1} = \vb_k, \beta \gets \min\{\frac{\beta}{t}, 1\}$.
		\ENDIF
		\ENDFOR
	\end{algorithmic}
\end{algorithm}

\subsection{Inexact APGnc}\label{sec: inex_APGnc}

We further consider inexact APGnc, in which computation of the proximal gradient step may be inexact, \ie,
  \[
  \xb_k = \prox{\eta g}^{\epsilon_k}(\yb_k - \eta (\nabla f(\yb_k) + \eb_k)),
  \]
where $\eb_k$ captures the inexactness of computation of $\nabla f(\yb_k)$, and $\epsilon_k$ captures the inexactness of evaluation of the proximal map as given by
\begin{align}
\xb &= \prox{\eta g}^{\epsilon}(\yb) \nonumber \\
&= \{\ub~|~ g(\ub) + \tfrac{1}{2\eta}\|\ub- \yb\|^2 \nonumber\\
 &\qquad \le \epsilon + g(\vb) + \tfrac{1}{2\eta}\|\vb- \yb\|^2,\quad \forall \vb\in \RR^d \}.
\end{align}
%These notions of inexactness have been considered in the literature as a standard setting \cite{}.
The inexact proximal algorithm has been studied in \cite{attouch_descent_semi} for nonconvex functions under the \KL property. Our study here is the first treatment of inexact proximal algorithms with momentum (i.e., APG-like algorithms). Furthermore, previous studies addressed only the inexactness of gradient computation for nonconvex problems, but our study here also includes the inexactness of the proximal map for nonconvex problems requiring only $g$ to be convex as the second case we specify below.

We study the following two cases.
\begin{enumerate}[leftmargin=*,topsep=0pt,noitemsep]
	\item $g$ is convex;
	\item $g$ is nonconvex, and $\epsilon = 0$.
\end{enumerate}
In the first case, $\partial g(\xb)$ reduces to the usual subdifferential of convex functions, and the inexactness $\epsilon$ naturally induces the following $\epsilon$-subdifferential
\begin{align}
	\partial_{\epsilon} g(\xb) = \{\ub~|~ g(\yb) \ge g(\xb) + \inner{\yb-\xb}{\ub} - \epsilon, \forall \yb \in\RR^d \}.\nonumber
\end{align}
Moreover, since the \KL property utilizes the information of $\partial F$,  we then need to characterize the perturbation of $\partial g$ under the inexactness $\epsilon$. This leads to the following definition.
\begin{definition}
	For any $\xb \in \RR^d$, let $\ub' \in \partial_{\epsilon} g(\xb)$ such that $\nabla f(\xb) + \ub'$ has the minimal norm. Then the perturbation between $\partial{g}$ and $\partial_{\epsilon} g$ is defined as $\xi: = \dist_{\partial g(\xb)} (\ub')$.
\end{definition}
%In general, since the \KL property only holds in a local neighborhood, we need to impose controls on the inexactness $\|\eb_k\|, \epsilon_k$ and $\xi_k$. We then have the following result regarding inexact APGnc.
The following theorem states that for nonconvex functions, as long as the inexactness parameters $\eb_k, \epsilon_k$ and $\xi_k$ are properly controlled, then the inexact APGnc converges at the same order-level rate as the corresponding exact algorithm.
\begin{thm}\label{thm: APGnc_inexact}
	Consider the above two cases for inexact APGnc under \Cref{assum:func}.\{1,2,3\}. If for all $k\in \mathbb{N}$
	\begin{align}
		\|\eb_k\| &\le \gamma \|\xb_k - \yb_k\|, \\
		 \epsilon_k &\le \delta \|\xb_k - \yb_k\|^2,\\
		 \xi_k &\le \lambda \|\xb_k - \yb_k\|,
	\end{align}
	then all the statements in \Cref{thm: limit_point} remain true and the convergence rates in \Cref{thm: KL_rate} remain at the same order with the constants $d_1 = (\frac{1}{\eta} + L + C)^2 /(\frac{1}{2\eta} - \frac{L}{2} - C)$, where $C>0$ depends on $\gamma, \delta$ and $\lambda$, and $d_2 = \min \{\frac{1}{2cd_1}, \frac{c}{1-2\theta}(2^{\frac{2\theta - 1}{2\theta - 2}} - 1)r_{k_0}^{2\theta - 1} \}$. Correspondingly, a smaller stepsize $\eta < \frac{1}{2C + L}$ should be used.
\end{thm}
It can be seen that, due to the inexactness, the constant factor $d_1$ in
\Cref{thm: KL_rate} is enlarged, which further leads to a smaller $d_2$ in
\Cref{thm: KL_rate}. Hence, the corresponding convergence rates are slower compared to the exact case, but remain at the same order.

%In general, one expects a slower convergence rate due to the inexactness, but it remains at the same order so long as the inexactness are well controlled.

\subsection{Stochastic Variance Reduced APGnc}
In this subsection, we study the stochastic variance reduced APGnc algorithm, referred to as SVRG-APGnc. The main steps are presented in \Cref{alg: svrg-APGnc}. The major difference from APGnc is that the single proximal gradient step is replaced by a loop of stochastic proximal gradient steps using variance reduced gradients.

Due to the stochastic nature of the algorithm, the iterate sequence may not stably stay in the local \KL region, and hence the standard \KL approach fails. We then focus on the analysis of the special but important case of the global \KL property with $\theta = \frac{1}{2}$. In fact, if $g=0$, the \KL property in such a case reduces to the well known Polyak-{\L}ojasiewicz (PL) inequality studied in \cite{Karimi2016}. Various nonconvex problems have been shown to satisfy this property such as quadratic phase retrieval loss function \cite{Yi2016} and neural network loss function \cite{Ma2016}. The following theorem characterizes the convergence rate of SVRG-APGnc under the \KL property with $\theta=\frac{1}{2}$.

\begin{algorithm}[tb]
	\caption{SVRG-APGnc}
	\label{alg: svrg-APGnc}
	\begin{algorithmic}
		\STATE {\bfseries Input:} $\yb_0 = \xb_0^0, \beta_k = \frac{k}{k+3}, m, \eta < \frac{1}{2mL}$.
		\FOR{$k=0, 1, 2, \cdots$}
		\STATE $\xb_k^0=\yb_k, \gb_k = \nabla f(\yb_k)$.
		\FOR{$t=0,1,\cdots,m-1$}
		\STATE sample $\xi$ from $\{1,2,\cdots,n\}$.
		\STATE $\vb_k^t=\nabla f_\xi (\xb_k^t)-\nabla f_\xi(\yb_k)+\gb_k$.
		\STATE $\xb^{t+1}_k= \prox{\eta g} (\xb_k^t-\eta \vb_k^t)$.
		\ENDFOR
		\STATE $\zb_k = \xb_k^m+\beta_k(\xb_k^m-\xb_{k-1}^m)$.
		\IF{$F(\xb_k^m) \le F(\zb_k)$}
		\STATE $\yb_{k+1} = \xb_k^m,$
		\ELSIF{$F(\zb_k) \le F(\xb_k^m)$}
		\STATE $\yb_{k+1} = \zb_k$.
		\ENDIF
		\ENDFOR
	\end{algorithmic}
\end{algorithm}

\begin{thm}\label{thm: svrg}
	Let $\eta=\rho/L$, where $\rho <1/2$ and satisfies $4\rho^2m^2+\rho \le 1.$	If the problem \textbf{(P)} satisfies the \KL property globally with $\theta=1/2$, then the sequence $\{\yb_k\}$ generated by \Cref{alg: svrg-APGnc} satisfies
	\begin{align}
	\Ebb\left[ F(\yb_k)-F^\ast\right] \le  \left(\tfrac{d}{d+1}\right)^{k}\left(F(\yb_0)-F^\ast\right),
	\end{align}
	where $d=\frac{c^2\left(L+\frac{1}{\eta}\right)^2+{\eta L^2 m}}{\frac{1}{2\eta}-L}$, and $F^\ast$ is the optimal function value.
\end{thm}
Hence, SVRG-APGnc also achieves the linear convergence rate under the \KL property with $\theta = \frac{1}{2}$. We note that \Cref{thm: svrg} differs from the linear convergence result established in \cite{NIPS2016_6116} for the SVRG proximal gradient in two folds: (1) we analyze proximal gradient with momentum but \cite{NIPS2016_6116}  studied proximal gradient algorithm; (2) the \KL property with $\theta = \frac{1}{2}$ here is different from the generalized PL inequality for composite functions adopted by \cite{Karimi2016}. In order to exploit the \KL property, our analysis of the convergence rate requires novel treatments of bounds, which can be seen in the proof of \Cref{thm: svrg} in \Cref{sec: thm: svrg}.

\subsection{Inexact SVRG-APGnc}
We further study the inexact SVRG-APGnc algorithm, and the setting of inexactness is the same as that in \Cref{sec: inex_APGnc}. Here, we focus on the case where $g$ is convex and $\eb_k=0$. The following theorem characterizes the convergence rate under such an inexact case.
%The more general cases are left to the future work.
\begin{thm}\label{thm: svrg_inexact}
	Let $g$ be convex and consider only the inexactness $\epsilon$ in the proximal map. Assume the \KL property is globally satisfied with $\theta=1/2$. Set $\eta=\rho/L$ where $\rho <1/2$ and satisfies 	$8\rho^2m^2+\rho \le 1.$ Assume that $\sum\limits_{t=0}^{m-1}3E\left[  \epsilon_k^t\right] \le \alpha \sum\limits^{m-1}_{t=0}E\left[\|\bar{\xb}_k^{t+1}-{\xb}_k^t\|^2 \right]$ for some $\alpha >0$, and define $\bar{\xb}^{t+1}_k=\prox{\eta g}(\xb^t_k-\eta\nabla f(\xb^t_k))$.
	Then the sequence $\{\yb_k\}$  satisfies
	\begin{align}
	\Ebb\left[ F(\yb_k)-F^\ast\right] \le  \left(\tfrac{d}{d+1}\right)^{k}\left(F(\yb_0)-F^\ast\right),
	\end{align}
	where $d=\frac{C^2\left(L+\frac{1}{\eta}\right)^2+{2\eta L^2 m+\frac{1}{2\eta}}}{\frac{1}{2\eta}-L-\alpha}$, and $F^\ast$ is the optimal function value.
\end{thm}
The convergence analysis for stochastic methods in inexact case has never been addressed before. To incorporate the \KL property in deriving the convergence rate, we use a reference sequence generated by exact proximal mapping. Even though this sequence is not actually generated by the algorithm, we can reach to the convergence rate by analyzing the relation between the reference sequence and the actual sequence generated by the algorithm.

Compared to the exact case, the convergence rate remains at the same order, i.e., the linear convergence, but the convergence is slower due to the larger parameter $d$ caused by the error parameter $\alpha$.  

\section{Experiments}\label{sec: exp}
In this section, we compare the efficiency of APGnc and SVRG-APGnc with other competitive methods via numerical experiments. In particular, we focus on the non-negative principle component analysis (NN-PCA) problem, which can be formulated as
\begin{align}
\min\limits_{\xb\ge 0} -\frac{1}{2}\xb^T \left(\sum\limits_{i=1}^n \mathbf{z}_i\mathbf{z}_i^T\right)\xb+\gamma\|\xb\|^2.%+\mathcal{I}_C
%=\min\limits_{\xb} \sum\limits_{i=1}^n f_i(\xb)+\mathcal{I}_C(\xb),
\end{align}
It can be equivalently written as
\begin{align}
	\min\limits_{\xb} -\frac{1}{2}\xb^T \left(\sum\limits_{i=1}^n \mathbf{z}_i\mathbf{z}_i^T\right)\xb+\gamma\|\xb\|^2+\mathbf{1}_{\{\xb\ge0 \}}.
\end{align}
Here, $f$ corresponds to the first two terms, and $g$ is the indicator of the nonnegative orthant, \ie, $\mathbf{1}_{\{\xb\ge0 \}}$.
This problem is nonconvex due to the negative sign and satisfies \Cref{assum:func}. In particular, it satisfies the \KL property since it is quadratic.

For the experiment, we set $n=2000, \gamma = 10^{-3}$ and randomly generate the samples $\zb_i$ from normal distribution. All samples are then normalized to have unit norm. The initialization is randomly generated, and is applied to all the methods.
We then compare the function values versus the number of effective passes through $n$ samples.
\begin{figure}[htp]
	\centering
	\begin{minipage}[t]{0.5\linewidth}
		\includegraphics[width=\linewidth]{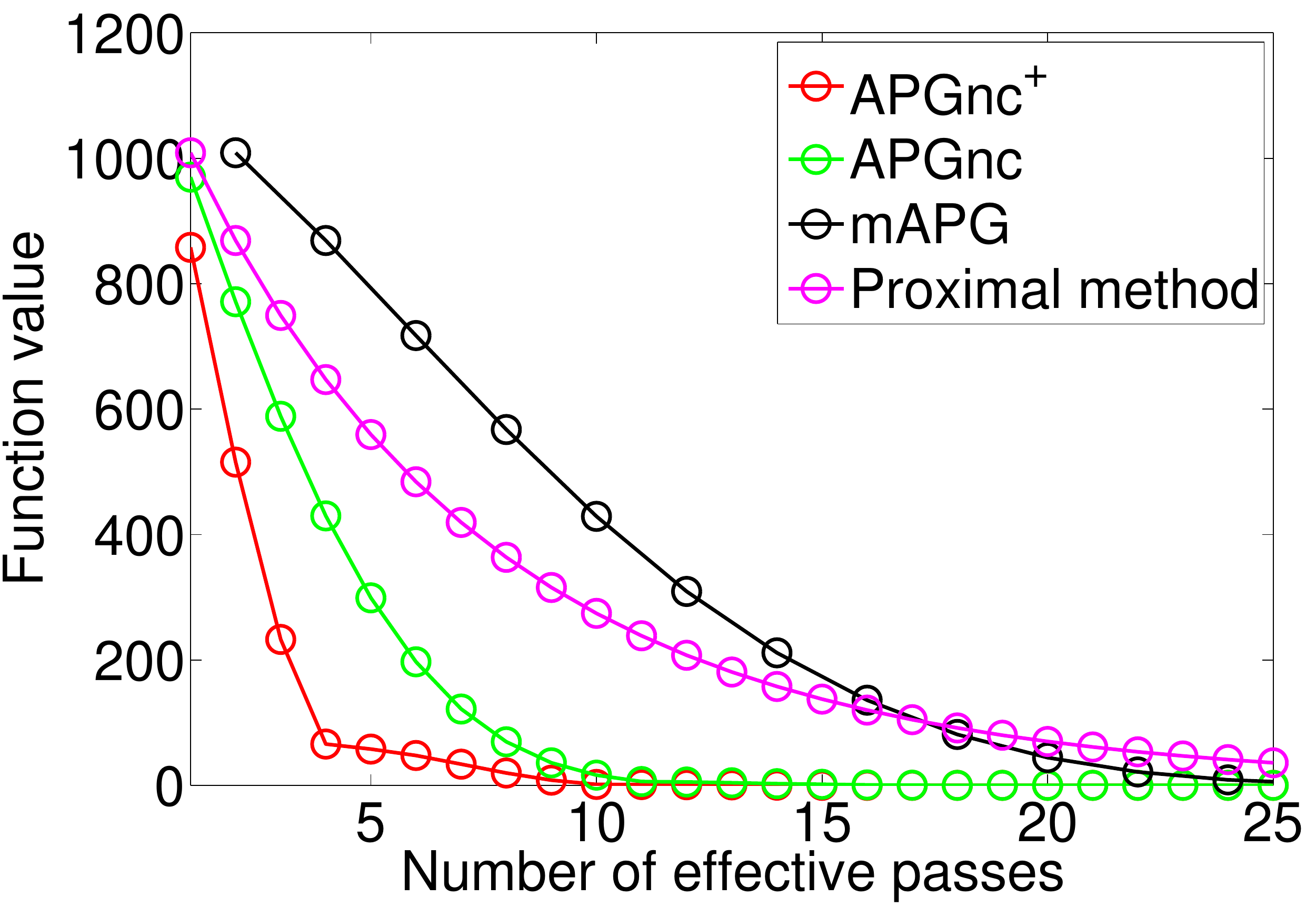}
		\subcaption{}
	\end{minipage}%
	\begin{minipage}[t]{0.5\linewidth}
		\includegraphics[width=\linewidth]{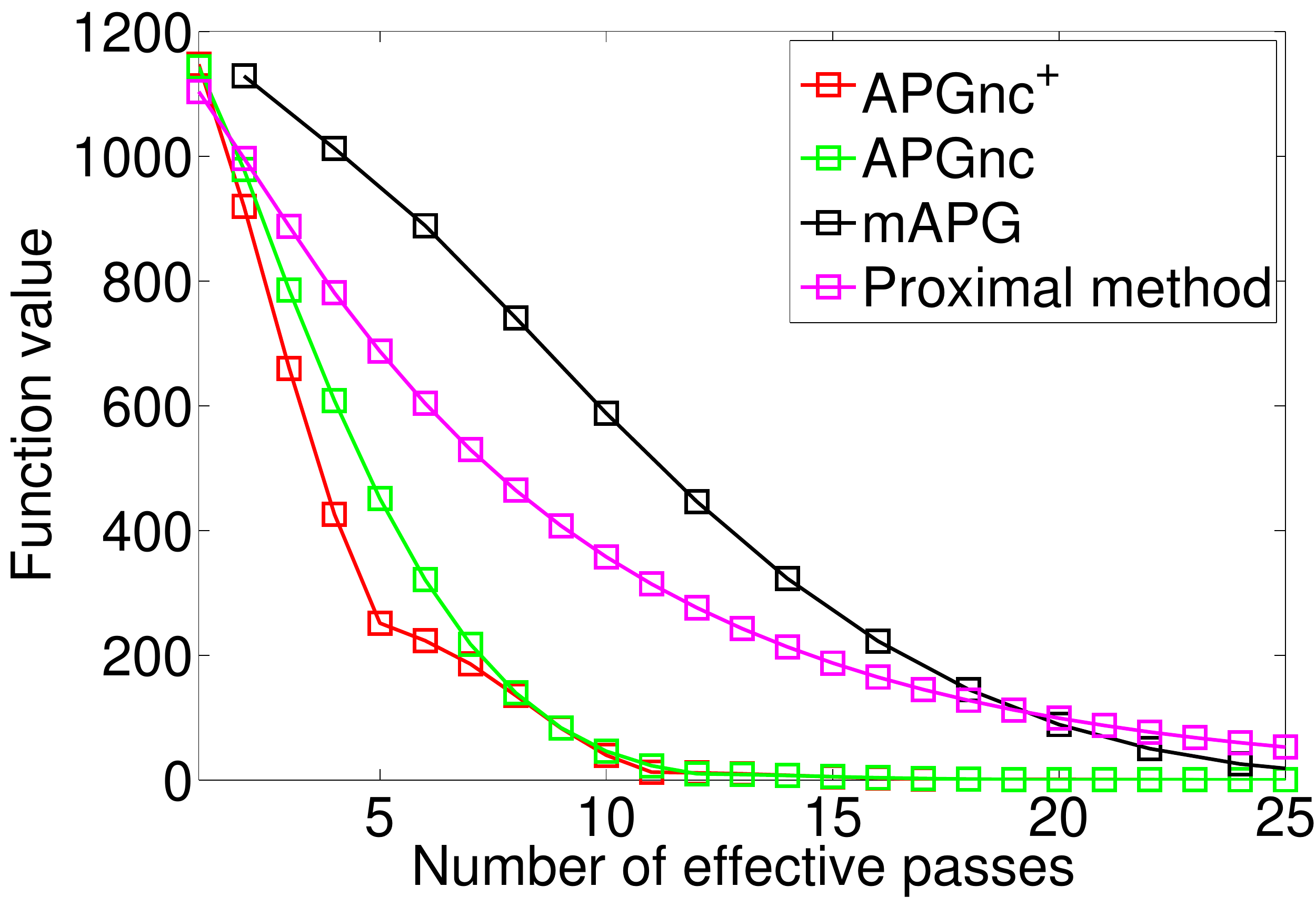}
		\subcaption{}
	\end{minipage}
	\caption{ Performance comparison of APGnc+, APGnc, mAPG, and the traditional proximal method. (a) Error free. (b) There exists the proximal error.}
\end{figure}
\begin{figure}[htp]
	\centering
	\begin{minipage}[t]{0.5\linewidth}
		\includegraphics[width=\linewidth]{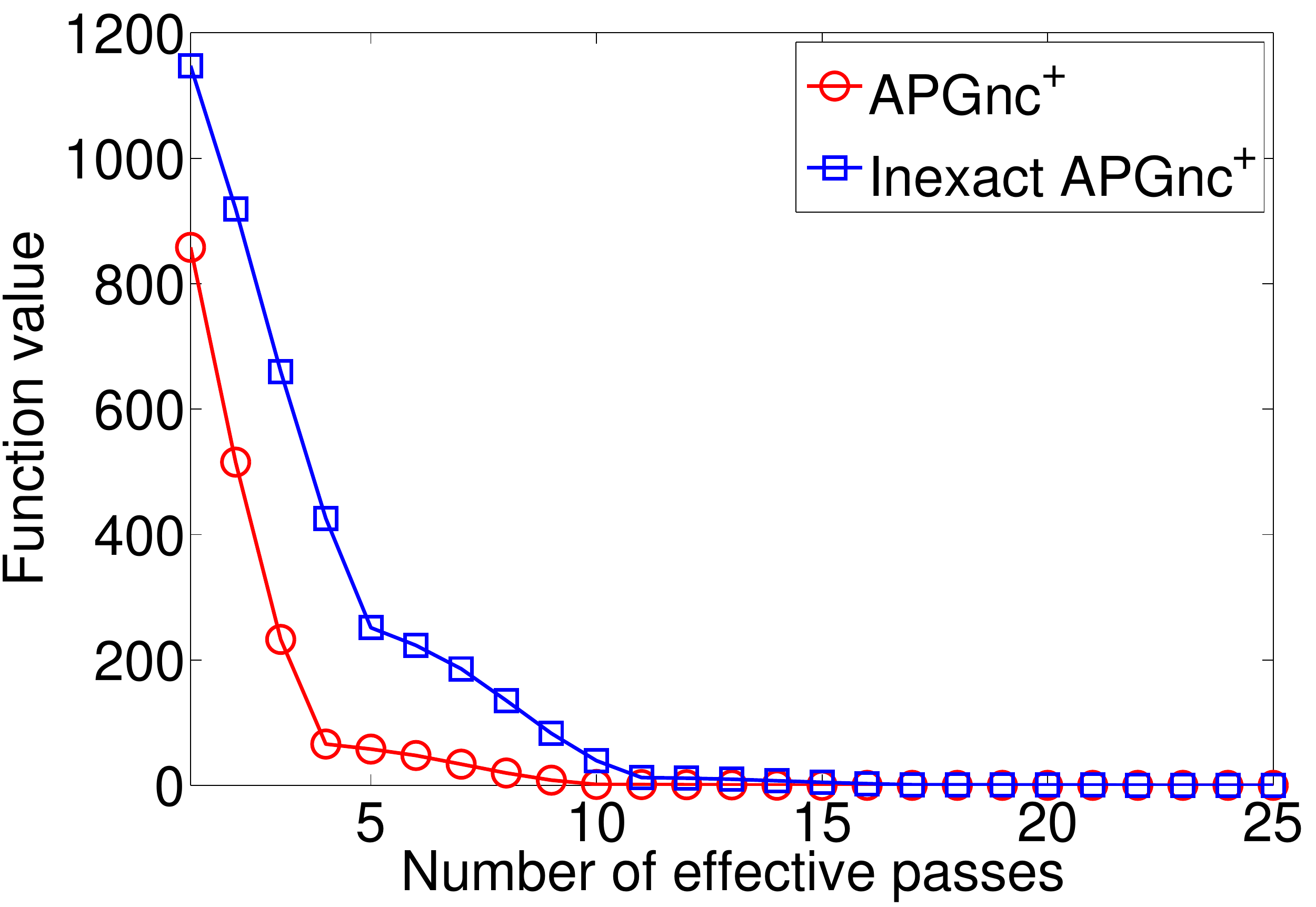}
		\subcaption{}
	\end{minipage}%
	\begin{minipage}[t]{0.5\linewidth}
		\includegraphics[width=\linewidth]{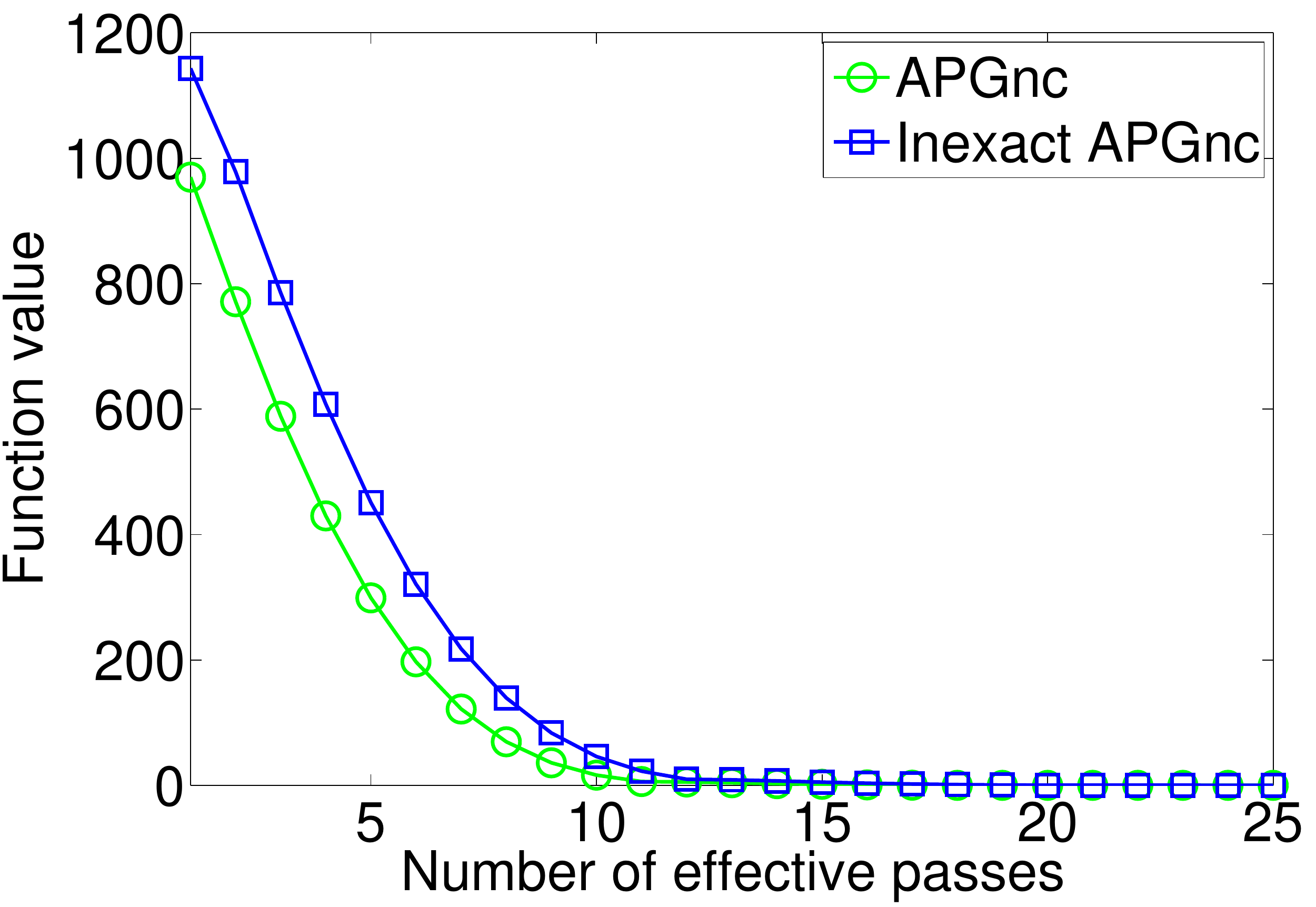}
		\subcaption{}
	\end{minipage}
	\caption{ Performance comparison of the same algorithm with and without the proximal error. (a) Performance comparison of APGnc+.  (b) Performance comparison of APGnc.}
\end{figure}
\subsection{Comparison among APG variants}
We first compare among the deterministic APG-like methods in Algorithms 2 - 4
and the standard proximal gradient method. The original APG in Algorithm 1 is not considered since it is not a descent method and does not have convergence guarantee in nonconvex cases. We use a fixed step size $\eta=0.05/L$, where $L$ is the spectral norm of the sample matrix $\sum\limits_{i=1}^n \mathbf{z}_i\mathbf{z}_i^T$. We set $t=1/2$ for APGnc$^+$. The results are shown in Figures 1 and 2. In Figure 1 (a), we show the performance comparison of the methods when there is no error in gradient or proximal calculation. One can see that APGnc and APGnc$^+$ outperform all other APG variants. In particular, APGnc$^+$ performs the best with our adaptive momentum strategy, justifying its empirical advantage. We note that the mAPG requires two passes over all samples at each iteration, and is, therefore, less data efficient compared to other APG variants.

We further study the inexact case in Figure 1 (b), where we introduce the proximal error $\epsilon_k=\frac{1}{100k^3}$ at the $k$th iteration. One can see that inexact APGnc$^+$ and inexact APGnc also outperform other two inexact algorithms. Furthermore, in Figure 2 (a) and (b), we compare exact and inexact algorithms respectively for APGnc$^+$ and APGnc. It can been that even with a reasonable amount of inexactness, both methods converge comparably to their corresponding exact methods. Although initially the function value drops faster in exact algorithms, both exact and inexact algorithms converge to the optimal point almost at the same time. Such a fact demonstrates the robustness of the algorithms.

\begin{figure}[htp]
	\centering
	\begin{minipage}[t]{0.5\linewidth}
		\includegraphics[width=\linewidth]{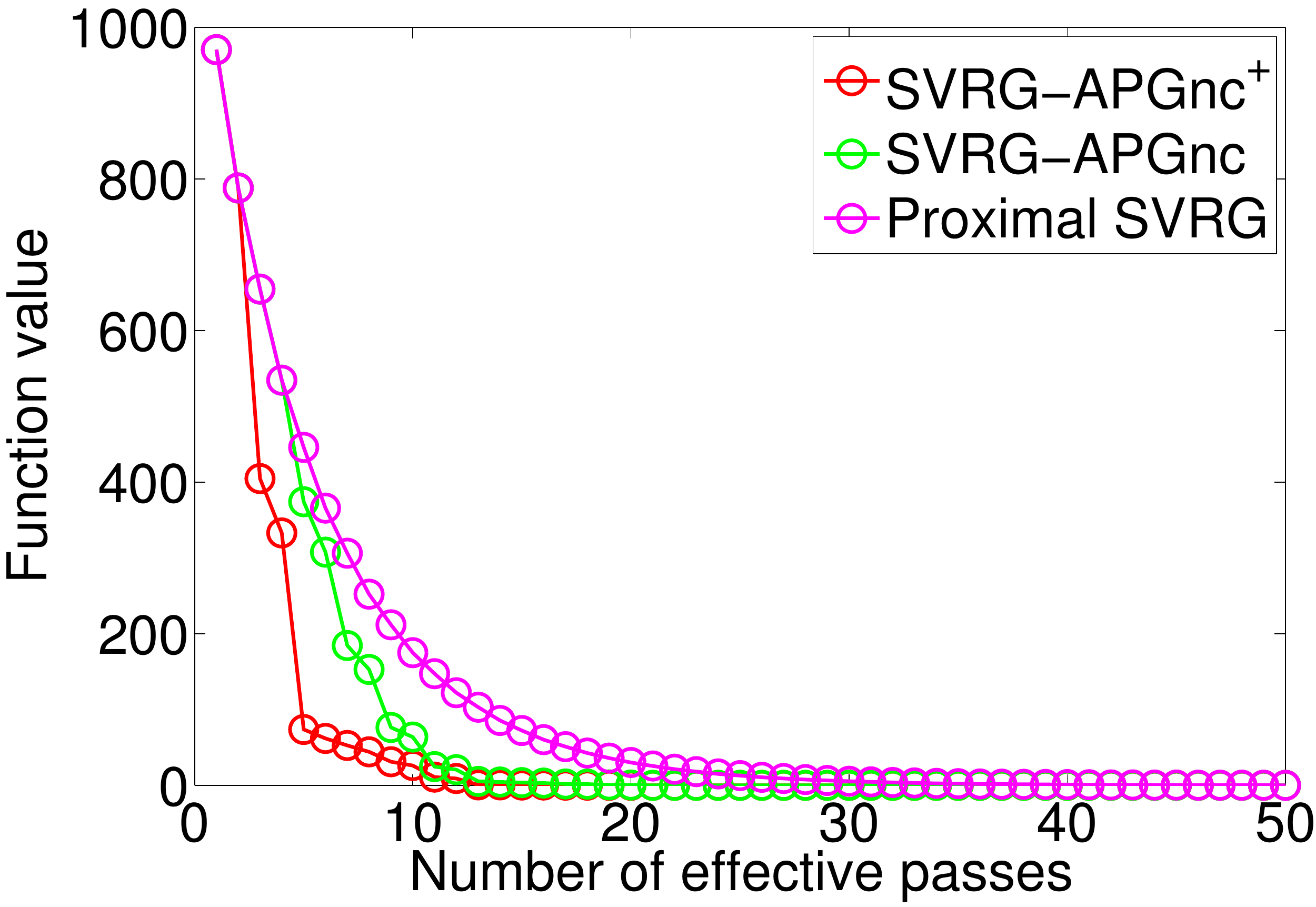}
		\subcaption{}
	\end{minipage}%
	\begin{minipage}[t]{0.5\linewidth}
		\includegraphics[width=\linewidth]{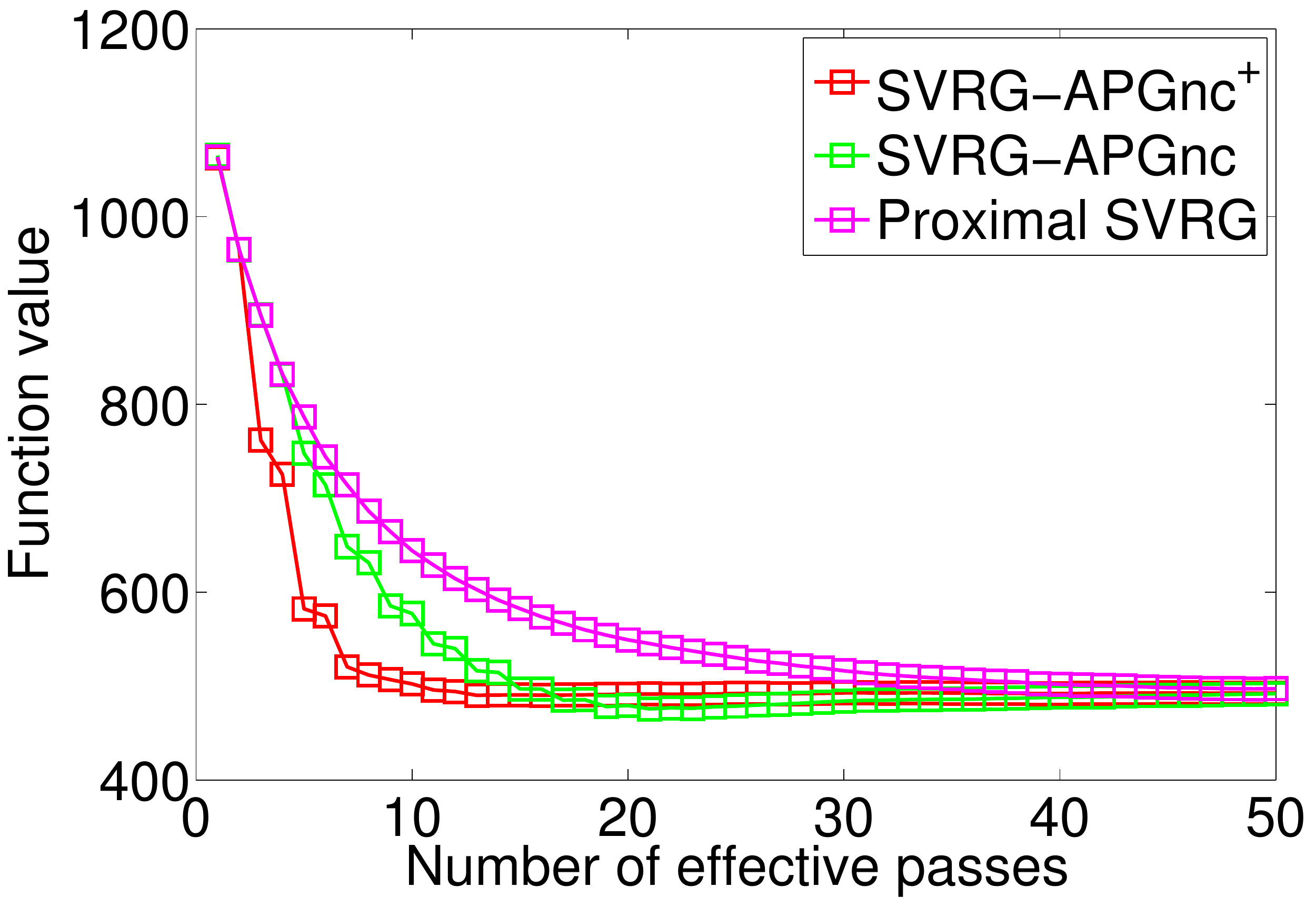}
		\subcaption{}
	\end{minipage}
	\caption{ Performance comparison of SVRG-APGnc+, SVRG-APGnc, and the traditional proximal SVRG. (a) Error free. (b) There exists the proximal error.}
\end{figure}
\begin{figure}[htp]
	\centering
	\begin{minipage}[t]{0.5\linewidth}
		\includegraphics[width=\linewidth]{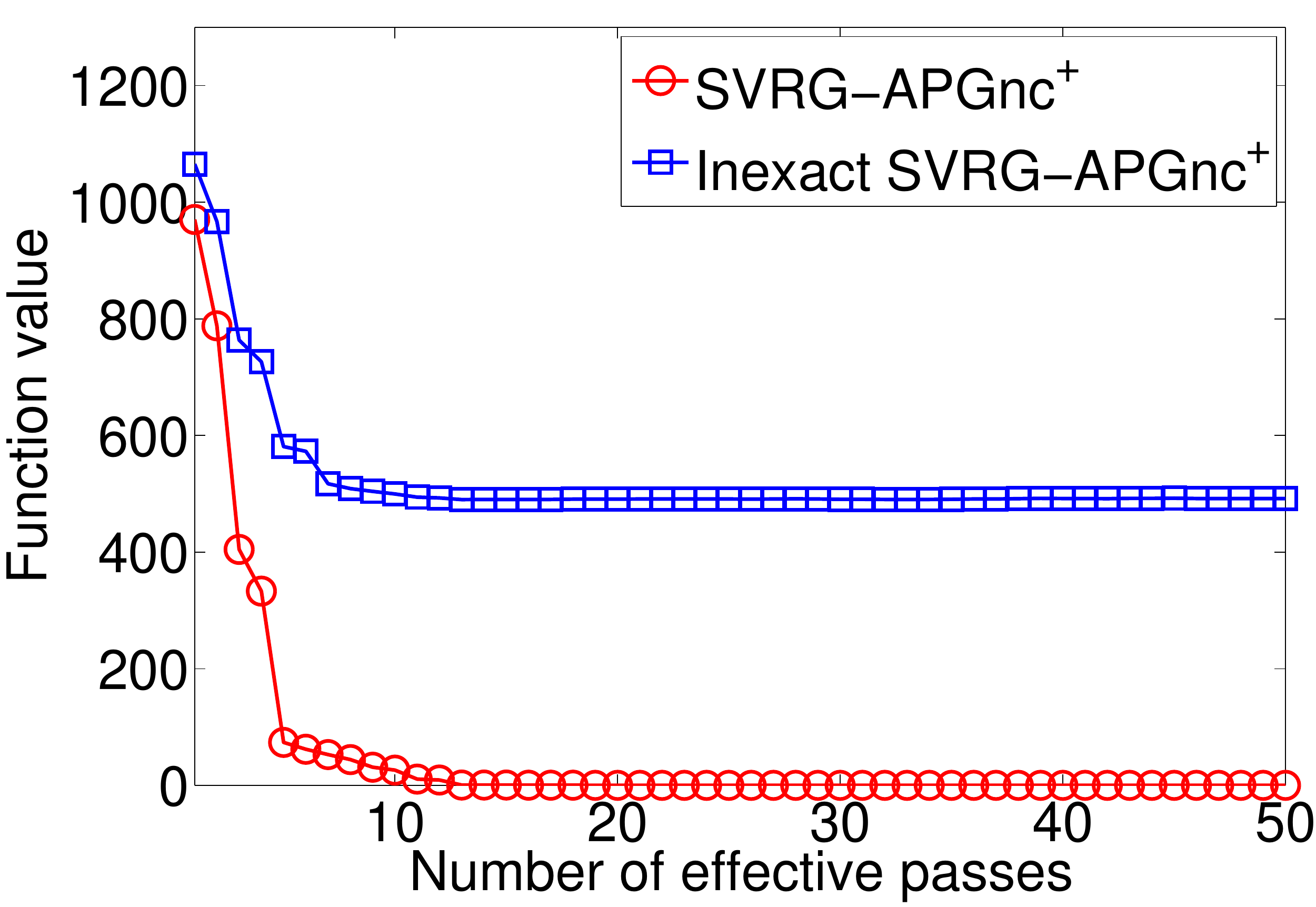}
		\subcaption{}
	\end{minipage}%
	\begin{minipage}[t]{0.5\linewidth}
		\includegraphics[width=\linewidth]{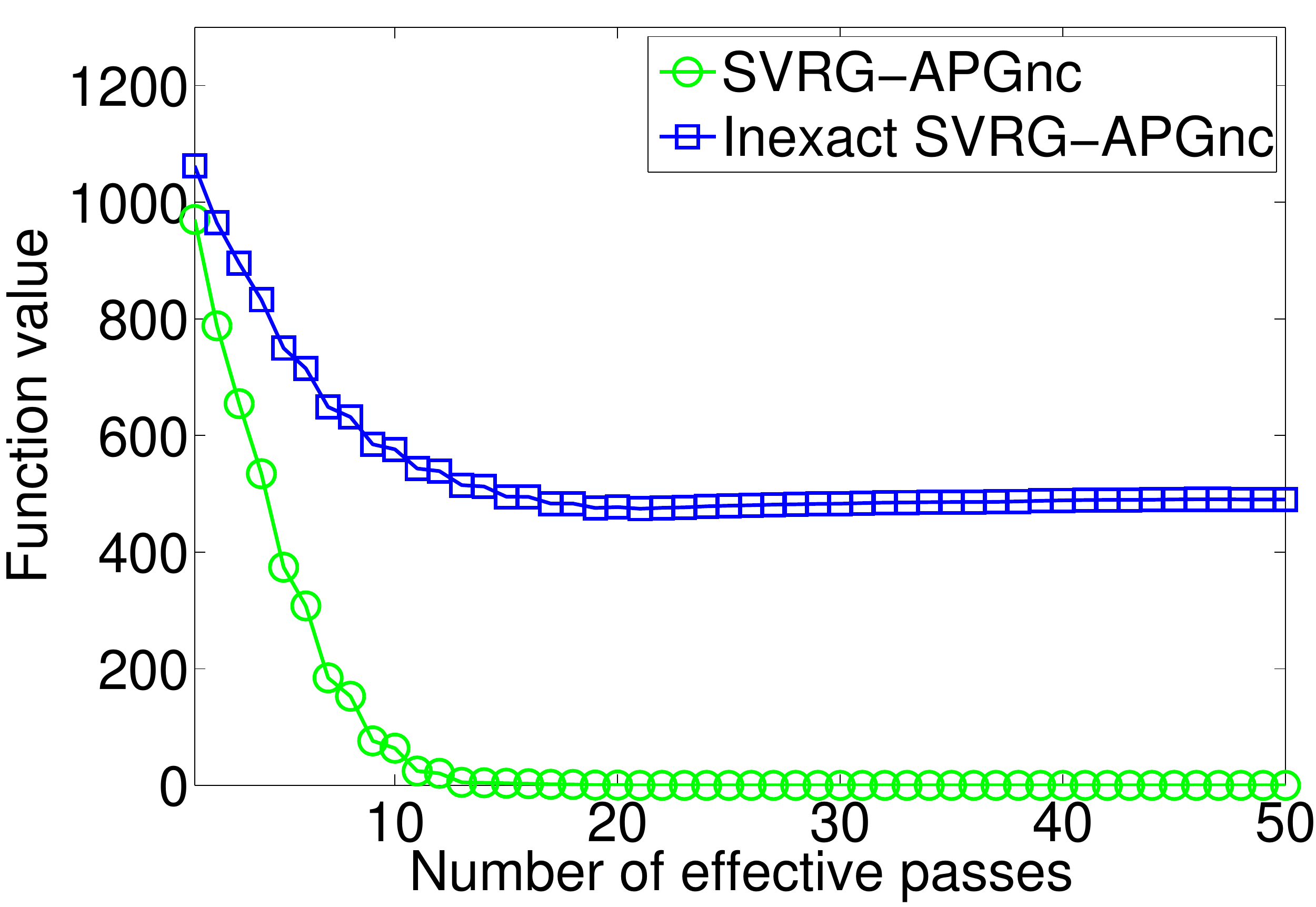}
		\subcaption{}
	\end{minipage}%
	\caption{ Performance comparison of the same algorithm with and without the proximal error. (a) Performance comparison of SVRG-APGnc+.  (b) Performance comparison of SVRG-APGnc.}
\end{figure}
\subsection{Comparison among SVRG-APG variants}
We then compare the performance among SVRG-APGnc, SVRG-APGnc$^+$ and the original proximal SVRG methods, and pick the stepsize $\eta=1/8mL$ with $m=n$.
The results are presented in Figures 3 and 4. In the error free case in Figure 3 (a), one can see that SVRG-AGPnc$^+$ method outperforms the others due to the adaptive momentum, and the SVRG-APGnc method also performs better than the original proximal SVRG method.

For the inexact case, we set the proximal error as $\epsilon_k=\min(\frac{1}{100k^3}, 10^{-7})$. One can see from Figure 3 (b) that the performance is degraded compared to the exact case, and converges to a different local minimum. In this result, all the methods are no longer monotone due to the inexactness and the stochastic nature of SVRG. Nevertheless, the SVRG-APGnc$^+$ still yields the best performance.

We also compare the results corresponding to SVRG-APGnc$^+$ and SVRG-APGnc, with and without the proximal error, in Figure 4 (a) and (b), respectively. It is clear that the SVRG-based algorithms are much more sensitive to the error comparing with APG-based ones. Even though the error is set to be smaller than in the inexact case with APG-based methods, one can observe more significant performance gaps than those in Figure 2.

\section{Conclusion}
In this paper, we provided comprehensive analysis of the convergence properties of APGnc as well as its inexact and stochastic variance reduced forms by exploiting the \KL property. We also proposed an improved algorithm APGnc$^+$ by adapting the momentum parameter. We showed that APGnc shares the same convergence guarantee and the same order of convergence rate as the mAPG, but is computationally more efficient and more amenable to adaptive momentum. In order to exploit the \KL property for accelerated algorithms in the situations with inexact errors and/or with stochastic variance reduced gradients, we developed novel convergence analysis techniques, which can be useful for exploring other algorithms for nonconvex problems.

%also explore the analysis of both inexact and SVRG variants of APGnc, and hope that all these variants can enrich the applicability of APG methods for solving general nonconvex problems.

\bibliography{./ref}
\bibliographystyle{icml2017}

\clearpage
\onecolumn
%\appendix
%
%\noindent {\Large \textbf{Supplementary Materials}}
%
%\input{appendix}

\end{document}